# WSEBP: A Novel Width-depth Synchronous Extension-based Basis Pursuit Algorithm for Multi-Layer Convolutional Sparse Coding

Haitong Tang, Shuang He, Lingbin Bian, Zhiming Cui, Nizhuan Wang*

*Abstract*—The pursuit algorithms integrated in multi-layer convolutional sparse coding (ML-CSC) can interpret the convolutional neural networks (CNNs). However, many current state-of-art (SOTA) pursuit algorithms require multiple iterations to optimize the solution of ML-CSC, which limits their applications to deeper CNNs due to high computational cost and large number of resources for getting very tiny gain of performance. In this study, we focus on the $0^{th}$ iteration in pursuit algorithm by introducing an effective initialization strategy for each layer, by which the solution for ML-CSC can be improved. Specifically, we first propose a novel width-depth synchronous extension-based basis pursuit (WSEBP) algorithm which solves the ML-CSC problem without the limitation of the number of iterations compared to the SOTA algorithms and maximizes the performance by an effective initialization in each layer. Then, we propose a simple and unified ML-CSC-based classification network (ML-CSC-Net) which consists of an ML-CSC-based feature encoder and a fully-connected layer to validate the performance of WSEBP on image classification task. The experimental results show that our proposed WSEBP outperforms SOTA algorithms in terms of accuracy and consumption resources. In addition, the WSEBP integrated in CNNs can improve the performance of deeper CNNs and make them interpretable. Finally, taking VGG as an example, we propose WSEBP-VGG13 to enhance the performance of VGG13, which achieves competitive results on four public datasets, i.e., 87.79% vs. 86.83% on Cifar-10 dataset, 58.01% vs. 54.60% on Cifar-100 dataset, 91.52% vs. 89.58% on COVID-19 dataset, and 99.88% vs. 99.78% on Crack dataset, respectively. The results show the effectiveness of the proposed WSEBP, the improved performance of ML-CSC with WSEBP, and interpretation of the CNNs or deeper CNNs.

*Index Terms*—Multi-layer convolutional sparse coding, Convolutional neural networks, Basis pursuit, VGG, Interpretability.

This work was supported by National Natural Science Foundation of China (No.61701318), Project of "Six Talent Peaks" of Jiangsu Province (No. SWYY-017), Project of Huaguoshan Mountain Talent Plan - Doctors for Innovation and Entrepreneurship.

Haitong Tang is with School of Biomedical Engineering, ShanghaiTech University and School of Marine Technology and Geomatics, Jiangsu Ocean University, China (e-mail: httang1224@gmail.com)
Shuang He is with School of Marine Technology and Geomatics, Jiangsu Ocean University, China (e-mail: kyrohe95@gmail.com).
Lingbing Bian, Zhiming Cui and Nizhuan Wang (corresponding author) are with School of Biomedical Engineering, ShanghaiTech University, China (e-mails: bianlb@shanghaitech.edu.cn; cuizm.neu.edu@gmail.com; wangnizhuan1120@gmail.com).

## I. INTRODUCTION

Convolutional neural networks (CNNs) have been tremendously successful in the last decade. Compared to traditional machine learning methods [1]–[3] extracting features manually, CNNs automatically extract features from input images through continuous learning of the convolutional layers, which are often considered as "black boxes" [4], [5]. The powerful capabilities of feature extraction using CNNs have been extensively applied to many tasks, including image classification [6]–[9], image segmentation [10]–[15], object detection [16]–[23], and so on [24]–[26].

Convolutional sparse coding (CSC) [27], one of the most popular models in the field of signal and image processing, has been proved with high feasibility and effectiveness in a wide range of applications [28]–[31]. In the framework of CSC, the signal is represented as a sparse linear combination of the union of cyclic and banded dictionaries [32], [33], by which the dictionaries are encoded from a global perspective with invariant shift and low redundancy. Further, the multi-layer convolutional sparse coding (ML-CSC) [34] is proposed to gain significant attention for its interpretability in deep learning [33]–[37].

Many pursuit algorithms have been proposed to solve representation vectors in ML-CSC, and the simplest one is the layered thresholding algorithm (LTA) [38] which solves the representation vector by convolution operations and provides a new interpretation of CNNs from the point of view of sparse representation: the forward pass process of CNNs is equivalent to the solving process of the encoding of ML-CSC. Meanwhile, Papyan et al. [38] have proposed a layered basis pursuit (LBP), an improved version of LTA, which considers each layer as a problem of independent basis pursuit and further improves the accuracy by introducing a large number of iterations at each layer. However, the LBP does not consider the layer-to-layer relationship sufficiently. To solve this problem, Sulam et al. [39] have proposed the multi-layer iteration soft thresholding algorithm (ML-ISTA), and its fast version, multi-layer fast iteration soft thresholding algorithm (ML-FISTA).

Although the aforementioned pursuit algorithms, i.e., LBP, ML-ISTA, ML-FISTA, etc., can be used to solve the problem in ML-CSC and further to explain and improve CNNs, there are several limitations such as the insufficient accuracy, relatively slow convergence rate, lack of effectiveness for deeper network optimization [40]. Therefore, this paper proposes a novel width-depth synchronous extension-based basis pursuit (WSEBP) algorithm, which solves the limitations mentioned above by the efficient initialization of every layer in the ML-CSC model. The proposed WSEBP can effectively reduce the computational resources caused by the large number of iterations while improving the accuracy compared to previous pursuit algorithms. Our main contributions are as follows:

1) A novel WSEBP algorithm with an efficient initialization of each layer is proposed to solve the problem in ML-CSC.
2) A simple and unified ML-CSC-based classification network (ML-CSC-Net) is proposed to validate the WSEBP and SOTA algorithms for image classification.
3) A new WSEBP-VGG13 model is proposed to enhance the performance of deeper CNNs and to interpret VGG13 [8].

The remainder of this paper is summarized as follows. In Section II, we introduce ML-CSC and its pursuit algorithms. In Section III, we introduce the WSEBP algorithm and describe how it can be used to improve and interpret CNNs. In Section IV, we first demonstrate the performance of the WSEBP algorithm using four different datasets and further propose WSEBP combined with VGG network as an example to improve the performance of VGG13. The paper is concluded in Section V.

## II. RELATED WORKS

In this section, we will introduce the theory and pursuit algorithms of ML-CSC.

### A. From Sparse Coding to Multi-Layer Convolutional Sparse Coding

The sparse coding (SC) assumes that a signal $\mathbf{X} \in \mathbb{R}^N$ can be represented by a linear combination of some columns with respect to the dictionary $\mathbf{D} \in \mathbb{R}^{N \times H}$, $H \gg N$. The procedure for finding $\mathbf{\Gamma} \in \mathbb{R}^H$ is shown as:

$$\min_{\mathbf{\Gamma}} \|\mathbf{\Gamma}\|_0 \\ \text{s.t. } \mathbf{X} = \mathbf{D}\mathbf{\Gamma} \quad (1)$$

where $\mathbf{\Gamma}$ is a vector of the coefficient called the representation vector, and the 0-norm is the sparsity constraint of $\mathbf{\Gamma}$. As images are regarded as high-dimensional signals, SC first decomposes the signal into many small overlapping image chunks and then performs a sparse representation on each chunk. However, this process is slow and often results in the loss of some important properties, e.g., translation invariance in the image.

In convolutional sparse coding (CSC), the signal is represented as a convolutional sum of a set of dictionary filters and the corresponding sparse coefficients. For signal $\mathbf{X} \in \mathbb{R}^N$, the model of CSC can be formulated as:

$$\mathbf{X} = \sum_{j=1}^{J} \mathbf{d}_j * \mathbf{\gamma}_j, \quad (2)$$

where $*$ is the convolution operator, $\{\mathbf{d}_j \in \mathbb{R}^M\}_{j \in \{1,2,\cdots,J\}}$ are the filters and $\{\mathbf{\gamma}_j \in \mathbb{R}^N\}_{j \in \{1,2,\cdots,J\}}$ are the sparse coefficient vectors. Further, a convolution operation in Eq. (2) can be

**Algorithm 1** The proposed WSEBP for solving ML-CSC
**Input:**
$\mathbf{X}$ - Original signal.
$\{\mathbf{D}_i\}_{i=1}^{L}$ - A set of convolutional dictionaries.
ReLU - Rectified Linear Unit.
$\{\xi_i\}_{i=1}^{L}, \{\beta_i\}_{i=1}^{L}$ - Learnable parameters.
**Output:**
A set of representations $\{\mathbf{\Gamma}_i\}_{i=1}^{L}$
**Process:**
1: $\mathbf{\Gamma}_0 = \mathbf{X}$
2: for $i = 1:L$ do:
3: $\quad \mathbf{\Gamma}_i \leftarrow \text{ReLU}\left(\frac{1}{\beta_i}\mathbf{D}_i^T\mathbf{\Gamma}_{i-1} + \mathbf{X} - \frac{1}{\beta_i}\mathbf{D}_i^T\mathbf{D}_i\mathbf{X} + \xi_i\right)$
4: end for

written as a matrix product, and $J$ convolutions correspond to $J$ cyclic matrices $\mathbf{D}_j \in \mathbb{R}^{N \times N}$, and each matrix is banded. Then, the model of CSC can be rewritten as:

$$\mathbf{X} = \sum_{j=1}^{J} \mathbf{d}_j * \mathbf{\gamma}_j = (\mathbf{D}_1 \cdots \mathbf{D}_J) \begin{pmatrix} \mathbf{\gamma}_1 \\ \vdots \\ \mathbf{\gamma}_J \end{pmatrix} = \mathbf{D}\mathbf{\Gamma}, \quad (3)$$

where all the cyclic matrices are arranged in a row to form $\mathbf{D} \in \mathbb{R}^{N \times JN}$ and all the sparse vectors are placed in a column to obtain the representation vector $\mathbf{\Gamma} \in \mathbb{R}^{JN}$. Thus, we can see that CSC is a special form of SC, whose dictionary is the union of cyclic and banded matrices.

Multi-layer convolutional sparse coding (ML-CSC) is an augmentation of CSC, which assumes that the representation vector $\mathbf{\Gamma}$ can further be expanded by performing CSC to form a cascaded multi-layer format as shown in Eq. (4).

$$\begin{aligned} \mathbf{X} &= \mathbf{D}_1 \mathbf{\Gamma}_1 \\ \mathbf{\Gamma}_1 &= \mathbf{D}_2 \mathbf{\Gamma}_2 \\ \mathbf{\Gamma}_2 &= \mathbf{D}_3 \mathbf{\Gamma}_3 \\ &\vdots \\ \mathbf{\Gamma}_{L-1} &= \mathbf{D}_L \mathbf{\Gamma}_L \end{aligned} \quad (4)$$

### B. Pursuit Algorithms of Multi-Layer Convolutional Sparse Coding

The solution for representation vectors $\{\mathbf{\Gamma}_i\}_{i=1}^{L}$ in Eq. (4) can be viewed as an optimization problem as follows:

$$\left\{ \min_{\mathbf{\Gamma}_i} \frac{1}{2} \|\mathbf{\Gamma}_{i-1} - \mathbf{D}_i \mathbf{\Gamma}_i\|_2^2 + \lambda_i \|\mathbf{\Gamma}_i\|_0 \right\}_{i=1}^{L}, \quad (5)$$

where the 2-norm indicates the accuracy of the solution. $\{\lambda_i\}_{i=1}^{L}$ is the regular term to balance accuracy and 0-norm enforces the sparsity. However, it is well-known that Eq. is an NP-hard problem[40]. It has been proved in [41] that Eq. (5) can be replaced with Eq. (6) by deflating 0-norm to 1-norm:

$$\left\{ \min_{\mathbf{\Gamma}_i} \frac{1}{2} \|\mathbf{\Gamma}_{i-1} - \mathbf{D}_i \mathbf{\Gamma}_i\|_2^2 + \lambda_i \|\mathbf{\Gamma}_i\|_1 \right\}_{i=1}^{L}. \quad (6)$$

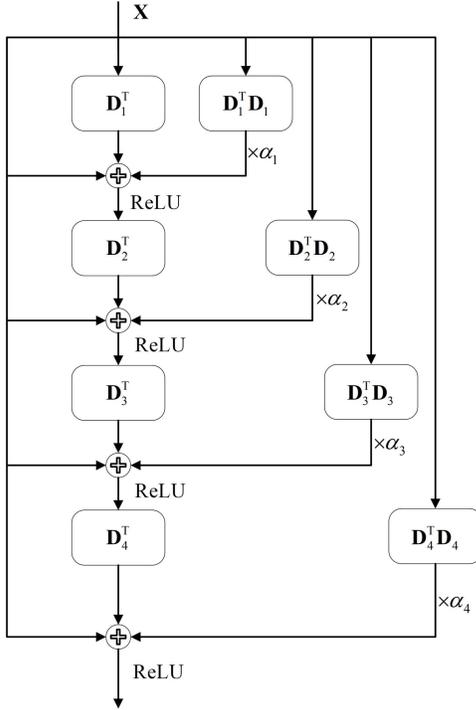

Fig. 1. The architecture of using WSEBP algorithm to solver four-layer ML-CSC. For convenience, we let $\alpha_i = 1/\beta_i$ and ignore the bias terms $\xi_i$.

*1) Layered thresholding algorithm*

Eq. can be solved by layered thresholding algorithm (LTA) [38], which performs the thresholding operation[42] independently for each layer to find the approximate solutions of representation vectors. The solving process is shown in Eq. :

$$\left\{ \mathbf{\Gamma}_i = S_{b_i}\left( \mathbf{D}_i^T \mathbf{\Gamma}_{i-1} \right) \right\}_{i=1}^{L}, \quad (7)$$

where $S_{b_i}$ is the threshold operator. It has been demonstrated in [38] that the threshold operator is equivalent to a translation with $b_i$ units followed by the Rectified Linear Unit (ReLU) function [43]. Thus, we can rewrite Eq. as follows

$$\begin{aligned}
\mathbf{\Gamma}_1 &= \text{ReLU}(\mathbf{D}_1^T \mathbf{X} + b_1) \\
\mathbf{\Gamma}_2 &= \text{ReLU}(\mathbf{D}_2^T \mathbf{\Gamma}_1 + b_2) \\
&\vdots \\
\mathbf{\Gamma}_L &= \text{ReLU}(\mathbf{D}_L^T \mathbf{\Gamma}_{L-1} + b_L)
\end{aligned} \quad (8)$$

*2) Layered Basis Pursuit Algorithm*

Eq. (6) also can be solved by the layered basis (LBP) algorithm [38] which uses the iterative soft thresholding algorithm (ISTA) [44] for each layer, and the solution is shown in Eq. :

$$\left\{ \mathbf{\Gamma}_i = \mathbf{\Gamma}_i^{k+1} = S_{\xi_i}\left( \mathbf{\Gamma}_i^k - \frac{1}{\beta_i} \mathbf{D}_i^T (\mathbf{D}_i \mathbf{\Gamma}_i^k - \mathbf{\Gamma}_{i-1}) \right) \right\}_{i=1}^{L}, \quad (9)$$

where $k$ is the number of iterations, $\mathbf{\Gamma}_0 = \mathbf{X}$, $\{\mathbf{\Gamma}_i^0\}_{i=1}^{L} = 0$, $\beta_i \geq \max\left(eigenvalue\left(\mathbf{D}_i^T \mathbf{D}_i\right)\right)$, and $\xi_i = \lambda_i / \beta_i$.

*3) Multi-layer Iteration Soft Thresholding Algorithm*

Sulam et al. [39] pointed out that the LBP algorithm only performs the ISTA for each layer separately, without considering the connection between each layer. To address this problem, they proposed the multi-layer iteration soft thresholding algorithm (ML-ISTA) to perform a global pursuit of whole layers and achieve better results. The solution via ML-ISTA is described in Eq. (10).

$$\left\{ \begin{aligned} \widehat{\mathbf{\Gamma}}_i &= \mathbf{D}_{(i,L)} \mathbf{\Gamma}_L^k \\ \mathbf{\Gamma}_i &= \mathbf{\Gamma}_i^{k+1} = S_{\xi_i}\left( \widehat{\mathbf{\Gamma}}_i - \frac{1}{\beta_i} \mathbf{D}_i^T (\mathbf{D}_i \widehat{\mathbf{\Gamma}}_i - \mathbf{\Gamma}_{i-1}) \right) \end{aligned} \right\}_{i=1}^{L}. \quad (10)$$

## III. METHOD

In this section, we first elaborate on the details of the proposed WSEBP; then, we propose the framework of WSEBP-VGG13, which shows a significant improvement compared to the deeper CNNs. The codes of our methods are available at: https://github.com/NZWANG/WSEBP.

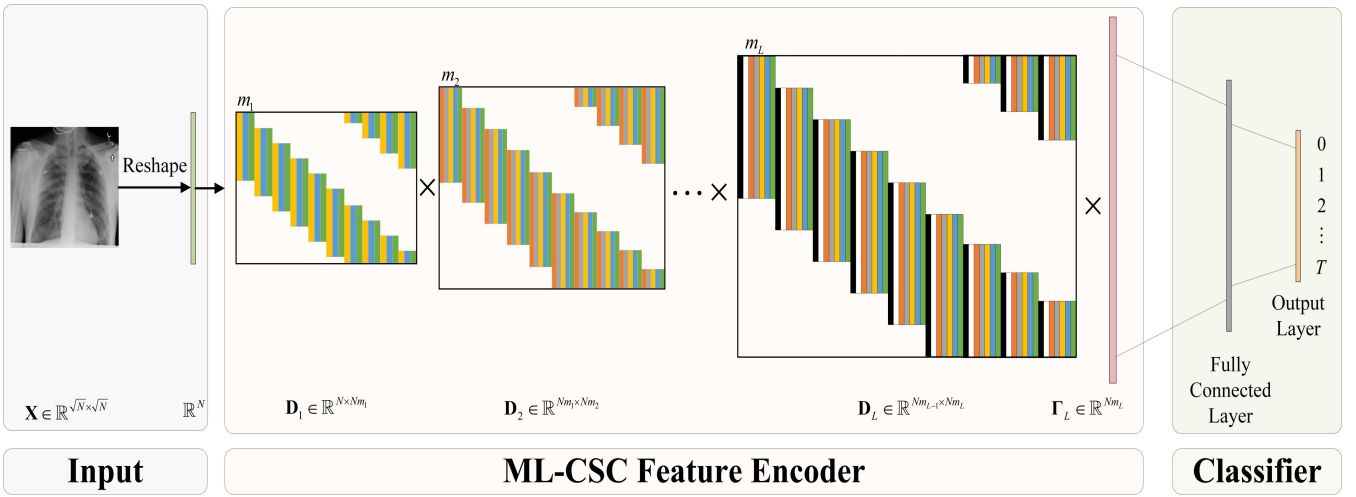

Fig. 2. The Framework of ML-CSC-Net which consist of ML-CSC block and fully connected layer. $\{m_i\}_{i=1}^{L}$ represent the number of filters in each layer respectively. $T+1$ is the category number of classification.

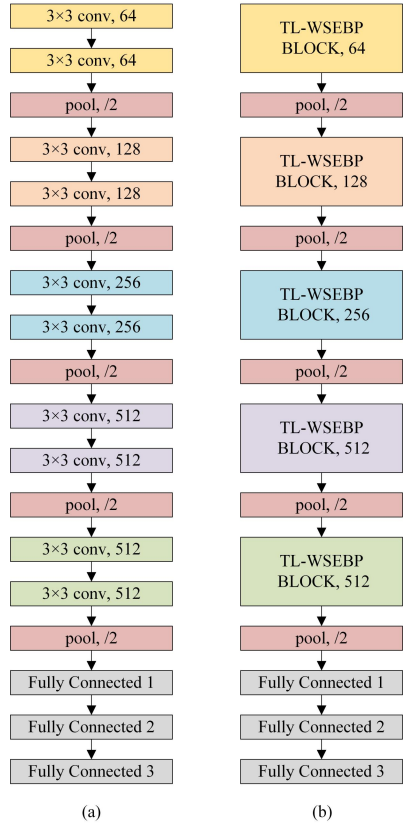

Fig. 3. The model structures of (a) VGG13 and (b) Our WSEBP-VGG13.

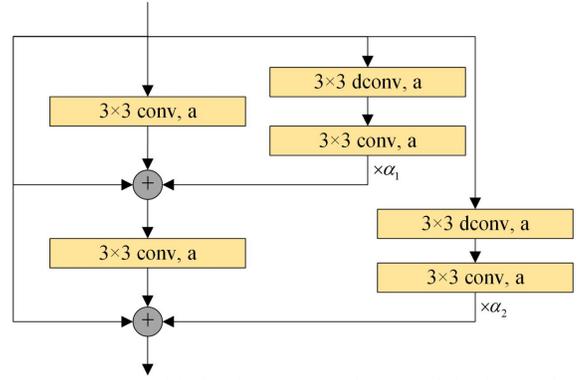

Fig. 4. TL-WSEBP block. The 3×3 convolution and the deconvolution operation represent $\mathbf{D}_i^T$ and $\mathbf{D}_i$, respectively.

### A. The Proposed WSEBP

In Section II, we have progressively introduced several pursuit algorithms for solving ML-CSC. The solutions using these algorithms are equivalent to each other if the number of iterations is zero. In comparison to LTA, LBP iteratively and continuously optimizes the $\{\mathbf{\Gamma}_i\}_{i=1}^L$ with dictionaries $\{\mathbf{D}_i\}_{i=1}^L$. In contrast to LBP, ML-ISTA uses a nested structure of multi-layer dictionaries and takes into account the links between each layer, such that the estimation accuracy of $\{\mathbf{\Gamma}_i\}_{i=1}^L$ is improved. Due to the difficulty for determining the most appropriate number of iterations in the pursuit algorithms, it is common practice to set a larger number of iterations to achieve more accurate results. However, a larger number of iterations will result in higher computational resource costs. In that case, the optimization algorithms set $\{\mathbf{\Gamma}_i^0\}_{i=1}^L = 0$ in the $0^{th}$ iteration for initialization whether using LBP or ML-ISTA. Inspired by the residual structure in ResNet [6], we set $\{\mathbf{\Gamma}_i^0\}_{i=1}^L = \mathbf{X}$ to accelerate the converge. Eq. in the $0^{th}$ iteration can be written as：

$$\left\{\mathbf{\Gamma}_i = \mathrm{ReLU}\left(\mathbf{X} - \frac{1}{\beta_i}\mathbf{D}_i^T(\mathbf{D}_i\mathbf{X} - \mathbf{\Gamma}_{i-1}) + \xi_i\right)\right\}_{i=1}^L, \quad (11)$$

which can be simplified by

$$\left\{\mathbf{\Gamma}_i = \mathrm{ReLU}\left(\frac{1}{\beta_i}\mathbf{D}_i^T\mathbf{\Gamma}_{i-1} + \mathbf{X} - \frac{1}{\beta_i}\mathbf{D}_i^T\mathbf{D}_i\mathbf{X} + \xi_i\right)\right\}_{i=1}^L \quad (12)$$

Based on Eq. (12), we propose a novel width-depth synchronous extension-based basis pursuit (WSEBP) algorithm, as presented in **Algorithm** 1. The core strategy in the WSEBP is taking advantage of $\{\mathbf{\Gamma}_i^0\}_{i=1}^L = \mathbf{X}$, which has superiority in speeding up iterative convergence of optimization by considering the relationships between each layer. Taking a four-layer ML-CSC as an example, the architecture of WSEBP is shown in Fig. 1, which uses forward propagation to update the representation vectors $\{\mathbf{\Gamma}_i\}_{i=1}^L$ and backward propagation to update the dictionaries $\{\mathbf{D}_i\}_{i=1}^L$, respectively.

### B. The ML-CSC-based Classification Network (ML-CSC-Net)

The ML-CSC can be treated as a feature extraction encoder, where the accuracy depends on the performance of pursuit algorithm used in the solver. The ML-CSC-based classification network (ML-CSC-Net) consists of a feature encoder of ML-CSC and a fully connected layer for classification (see in Fig. 2). This classification network can be used to evaluate the performance of pursuit algorithms, i.e., LTA, LBP, ML-ISTA, and WSEBP.

### C. Performance Improvement of VGG

VGG is a series of models such as VGG11, VGG 13, VGG16, and VGG19 [8]. It opens an era of small convolutional kernels in the large-scale visual recognition. VGG also has been used as a backbone network for a range of image tasks such as classification, localization, detection, and segmentation. In this study, taking the popularly used VGG13 as an example, we use the proposed WSEBP to enhance VGG13. The network structure of VGG13 is shown in Fig. 3a, which performs two-layer convolution operation followed by a max pooling operation. Based on this architecture, we construct a two-layer WSEBP, named TL-WSEBP block as shown in Fig. 4, to replace the two-layer convolution operation in VGG13, by which the WSEBP-VGG13 is obtained (see Fig. 3b). It is worth noting that the proposed WSEBP can be applied to any network involving convolutional operation.

TABLE I
THE DETAIL OF DATASETS USED IN THE EXPERIMENT

| Dataset | Class | Train | Validation | Test |
|---|---|---|---|---|
| Cifar-10 | 10 | 40,000 | 10,000 | 10,000 |
| Cifar-100 | 100 | 40,000 | 10,000 | 10,000 |
| COVID-19 | 4 | 12,698 | 4,233 | 4,235 |
| Crack | 2 | 24,000 | 8,000 | 8,000 |

TABLE II
THE PERFORMANCE OF CLASSIFICATION USING COMPETING PURSUIT ALGORITHMS ON CIFAR-10, CIFAR-100, COVID-19, AND CRACK DATASETS.

| Dataset | Algorithm | Layer | Parameter (M) | Accuracy (test)% | Memory Usage (M) |
|---|---|---|---|---|---|
| Cifar-10 | LTA | 4 | 0.178 | 75.50 | 716 |
| | LBP | 4 | 0.178 | 77.05 | 808 |
| | ML-ISTA | 4 | 0.178 | 76.77 | 810 |
| | WSEBP | 4 | 0.178 | **77.77** | **770** |
| Cifar-100 | LTA | 3 | 0.144 | 47.56 | 714 |
| | LBP | 3 | 0.144 | 47.89 | 802 |
| | ML-ISTA | 3 | 0.144 | 48.03 | 804 |
| | WSEBP | 3 | 0.144 | **48.78** | **766** |
| COVID-19 | LTA | 4 | 0.706 | 88.68 | 938 |
| | LBP | 4 | 0.706 | 89.44 | 1,662 |
| | ML-ISTA | 4 | 0.706 | 88.92 | 1,712 |
| | WSEBP | 4 | 0.706 | **90.53** | **1,356** |
| Crack | LTA | 3 | 0.014 | 99.22 | 812 |
| | LBP | 3 | 0.014 | 99.41 | 1,126 |
| | ML-ISTA | 3 | 0.014 | 99.55 | 1,188 |
| | WSEBP | 3 | 0.014 | **99.69** | **956** |

## IV. EXPERIMENT AND ANALYSIS

We first evaluate the performance of the WSEBP algorithm on four different datasets using the ML-CSC-based classification network, and then validate the performance of WSEBP-VGG13. The experiments are conducted on the Ubuntu operating system with one Nvidia GeForce GTX 1080Ti GPU, and the deep learning framework is implemented on the platform PyTorch [45].

### A. Datasets

The COVID-19[1] dataset is a public dataset with four types of chest X-ray (CXR) images: COVID-19 positive cases, normal, Lung Opacity and viral pneumonia. The Cifar-10 and Cifar-100 datasets are labeled as the subsets of the dataset with 80 million tiny images. These datasets are collected by Alex et.al [46] and commonly used to verify the performance of image classification models in the computer vision society. The Crack dataset [47] contains both negative and positive concrete crack images, and is generated from 458 high resolution METU campus building crack.

In previous related articles [37], [39], the datasets were divided into training and testing datasets, and the result on the testing set was directly used to evaluate the performance of the algorithm. In our experiment, we select a random part from the training set as a validation set which is used to select the best model, and then test the performance of competing models on the testing set. The details of the involved datasets are shown in Table I.

### B. Evaluating the Performance of WSEBP

#### 1) Competing algorithms

We compare our WSEBP with other popular pursuit algorithms, i.e., LTA, LBP, and ML-ISTA, under the ML-CSC-based classification network (ML-CSC-Net). The number of iterations of LBP and ML-ISTA are set to 2. The results of classification are used to judge the performance of the aforementioned pursuit algorithms.

#### 2) Experimental results

We evaluate the performance of WSEBP and compared to other state-of-the-art pursuit algorithms on the validation and testing datasets of four datasets. The specific experimental details of pursuit algorithms are presented in Appendix A. The comparison of the performance of our proposed method and other pursuit algorithms on the validation datasets are shown in Fig. 5, and the experimental results on the testing dataset are recorded in Table II.

From Fig. 5, we can see that the training performance on the validation datasets regarding to the accuracy using WSEBP outperforms all the other pursuit algorithms. According to Table II, our WSEBP achieves the best classification results on the testing datasets of Cifar-10 (i.e., 77.77%), Cifar-100 (i.e., 48.78%), COVID-19 (i.e., 90.53%), and Crack (i.e., 99.69%), respectively. Further, regarding the size of parameters, all the competing algorithms have the identical scale on the same dataset. Moreover, in terms of memory usage, our WSEBP consumes less GPU memory than other pursuit algorithms except the LTA. Overall, our proposed WSEBP achieves the highest accuracy while consumes less resource compared to the other pursuit algorithms.

### C. Evaluating the Performance of WSEBP-VGG13

In this section, we evaluate the WSEBP-VGG13 on four datasets. The classification results in testing dataset are shown in Table III and the details of experimental implementation of WSEBP-VGG13 are presented in Appendix B. From Table III, we can see that the proposed WSEBP-VGG13 outperforms the original VGG13 on all datasets. The proposed WSEBP can improve the performance of the model regarding to the convolutional operations. More importantly, it is a general

TABLE III
THE RESULTS OF CLASSIFICATION ON DIFFERENT DATASETS USING VGG AND WSEBP-VGG13.

| Model | Four Datasets | | | |
|---|---|---|---|---|
| | Cifar-10 | Cifar-100 | COVID-19 | Crack |
| VGG13 | 86.83 | 54.60 | 89.58 | 99.78 |
| WSEBP-VGG13 | **87.79** | **58.01** | **91.52** | **99.88** |

[1] https://www.kaggle.com/tawsifurrahman/covid19-radiography-database

method and can be directly applied to interpret the CNNs model or its variants.

## V. CONCLUSION

In this paper, we propose a novel WSEBP algorithm to solve ML-CSC, which learns the representation vector with an effective initialization of each layer instead of setting large number of iterations. Further, we propose a simple and unified ML-CSC-Net to validate the performance of WSEBP on image classification task. Meanwhile, we have showed theoretical superiority of WSEBP integrated in ML-CSC-Net. The extensive experiments demonstrate that our WSEBP provides a more accurate solution than other state-of-the-art pursuit algorithms, including LTA, LBP, and ML-CSC, while requiring less memory consumption. In addition, taking VGG13 as an example, we propose WSEBP-VGG13 by replacing convolution operation with WSEBP, by which CNNs are improved and interpreted. The experimental results show that WSEBP-VGG13 outperforms VGG13. Overall, our proposed WSEBP algorithm provides a new perspective for the optimization of ML-CSC and an alternative interpretation of CNNs.

## APPENDIX

### A. Experimental Details of Pursuit Algorithms

On the Cifar-10 dataset, we construct the ML-CSC-Net with four convolutional layers of size $4 \times 4$, with 16, 32, 64, and 128 convolution kernels, followed by a fully connected layer. The input image size is $32 \times 32$; the initial learning rate is 0.005; the momentum is 0.9; the batch size is 128; and a total of 200 epochs are trained, with the learning rate multiplied by 0.2 at 100 and 150 epochs.

On the Cifar-100 dataset, we construct the ML-CSC-Net with three convolution layers of size $4 \times 4$, with 16, 32 and 64 convolution kernels, followed by a fully connected layer. The input image size is $32 \times 32$; the initial learning rate is 0.005; the momentum is 0.9; the batch size is 128; a total of 200 epochs are trained, with the learning rate multiplied by 0.5 at 100 and 150 epochs.

On the COVID-19 dataset, we construct ML-CSC-Net with four convolution layers of size $4 \times 4$, with the number of convolution kernels being 32, 64, 128 and 256, followed by a fully connected layer. The input size is $64 \times 64$, the initial learning rate is 0.1, the momentum is 0.9, the batch size is 128, a total of 200 epochs are trained, and the learning rate is multiplied by 0.1 at 100 and 150 epochs.

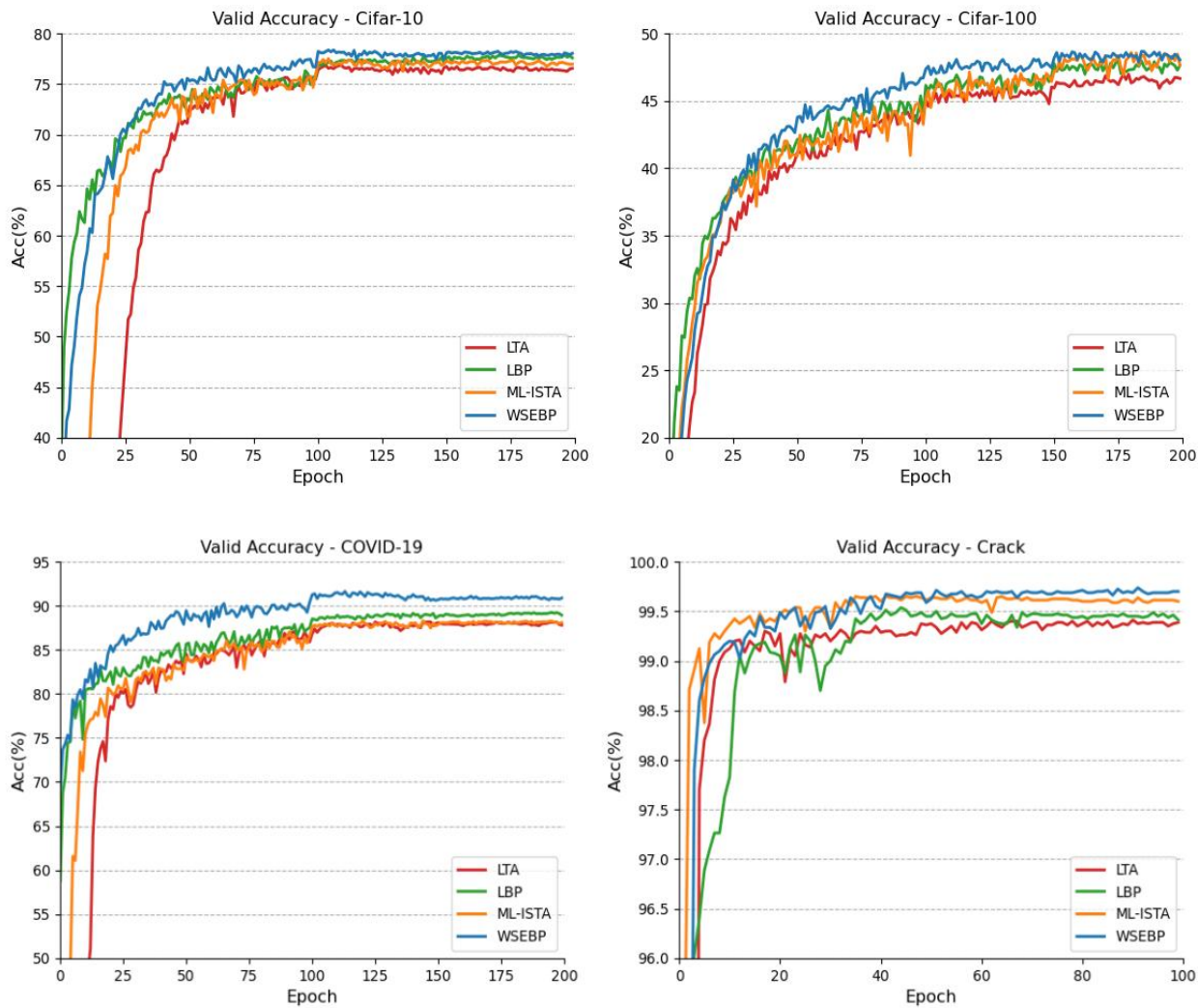

Fig. 5. Performance of competing algorithms on the validation datasets from Cifar-10, Cifar-100, COVID-19, and Crack.

On the Crack dataset, we construct ML-CSC-Net with three convolution layers of size 4×4, with 8, 16 and 32 convolution kernels in order, followed by a fully connected layer. The input size is 64 × 64, the initial learning rate is 0.01, the momentum is 0.9, and the batch size is 256. 100 epochs are trained, and the learning rate is multiplied by 0.5 at 40 and 70 epochs.

*B. Experimental Details of WSEBP-VGG13*

On the Cifar-10 dataset, the input size of WSEBP-VGG13 is 32×32, the initial learning rate is 0.01, the momentum is 0.9, the batch size is 128, and a total of 200 epochs are trained, with the learning rate multiplied by 0.1 at 100 and 150 epochs.

On the Cifar-100 dataset, the input size of WSEBP-VGG13 is 32 × 32, the initial learning rate is 0.005, the momentum is 0.9, the batch size is 128, and a total of 200 epochs are trained, with the learning rate multiplied by 0.5 at 100 and 150 epochs.

On the COVID-19 dataset, the input size of WSEBP-VGG13 is 64 × 64, the initial learning rate is 0.001, the momentum is 0.9, the batch size is 128, a total of 150 epochs are trained, and the learning rate is multiplied by 0.5 at 100 epochs.

On the crack dataset, the input size of WSEBP-VGG13 is 64×64, the initial learning rate is 0.001, the momentum is 0.9, and the batch size is 128, a total of 100 epochs are trained, and the learning rate is multiplied by 0.5 at 40 and 70 epochs.

REFERENCES

[1] A. Liaw and M. Wiener, "Classification and regression by randomForest," *R news*, vol. 2, no. 3, pp. 18–22, 2002.
[2] C. Cortes and V. Vapnik, "Support-vector networks," *Mach. Learn.*, vol. 20, no. 3, pp. 273–297, 1995.
[3] N. Dalal and B. Triggs, "Histograms of oriented gradients for human detection," in *2005 IEEE Computer Society Conference on Computer Vision and Pattern Recognition (CVPR'05)*, 2005, vol. 1, pp. 886–893 vol. 1, doi: 10.1109/CVPR.2005.177.
[4] J. M. Benítez, J. L. Castro, and I. Requena, "Are artificial neural networks black boxes?," *IEEE Trans. neural networks*, vol. 8, no. 5, pp. 1156–1164, 1997.
[5] V. Buhrmester, D. Münch, and M. Arens, "Analysis of explainers of black box deep neural networks for computer vision: A survey," *Mach. Learn. Knowl. Extr.*, vol. 3, no. 4, pp. 966–989, 2021.
[6] K. He, X. Zhang, S. Ren, and J. Sun, "Deep residual learning for image recognition," in *Proceedings of the IEEE Computer Society Conference on Computer Vision and Pattern Recognition*, 2016, vol. 2016-Decem, pp. 770–778, doi: 10.1109/CVPR.2016.90.
[7] A. Krizhevsky, I. Sutskever, and G. E. Hinton, "Imagenet classification with deep convolutional neural networks," *Adv. Neural Inf. Process. Syst.*, vol. 25, pp. 1097–1105, 2012.
[8] K. Simonyan and A. Zisserman, "Very deep convolutional networks for large-scale image recognition," *arXiv Prepr. arXiv1409.1556*, 2014.
[9] C. Szegedy *et al.*, "Going deeper with convolutions," *Proc. IEEE Comput. Soc. Conf. Comput. Vis. Pattern Recognit.*, vol. 07-12-June, pp. 1–9, 2015, doi: 10.1109/CVPR.2015.7298594.
[10] V. Badrinarayanan, A. Kendall, and R. Cipolla, "SegNet: A Deep Convolutional Encoder-Decoder Architecture for Image Segmentation," *IEEE Trans. Pattern Anal. Mach. Intell.*, vol. 39, no. 12, pp. 2481–2495, 2017, doi: 10.1109/TPAMI.2016.2644615.
[11] L. C. Chen, G. Papandreou, F. Schroff, and H. Adam, "Rethinking atrous convolution for semantic image segmentation," *arXiv*, 2017.
[12] L.-C. Chen, G. Papandreou, I. Kokkinos, K. Murphy, and A. L. Yuille, "Semantic image segmentation with deep convolutional nets and fully connected crfs," *arXiv Prepr. arXiv1412.7062*, 2014.
[13] L.-C. Chen, G. Papandreou, I. Kokkinos, K. Murphy, and A. L. Yuille, "Deeplab: Semantic image segmentation with deep convolutional nets, atrous convolution, and fully connected crfs," *IEEE Trans. Pattern Anal. Mach. Intell.*, vol. 40, no. 4, pp. 834–848, 2017.
[14] J. Long, E. Shelhamer, and T. Darrell, "Fully convolutional networks for semantic segmentation," in *Proceedings of the IEEE conference on computer vision and pattern recognition*, 2015, pp. 3431–3440.
[15] O. Ronneberger, P. Fischer, and T. Brox, "U-net: Convolutional networks for biomedical image segmentation," *Lect. Notes Comput. Sci. (including Subser. Lect. Notes Artif. Intell. Lect. Notes Bioinformatics)*, vol. 9351, pp. 234–241, 2015, doi: 10.1007/978-3-319-24574-4_28.
[16] S. Ren, K. He, R. Girshick, and J. Sun, "Faster r-cnn: Towards real-time object detection with region proposal networks," *arXiv Prepr. arXiv1506.01497*, 2015.
[17] R. Girshick, "Fast r-cnn," in *Proceedings of the IEEE international conference on computer vision*, 2015, pp. 1440–1448.
[18] J. Redmon, S. Divvala, R. Girshick, and A. Farhadi, "You only look once: Unified, real-time object detection," *Proc. IEEE Comput. Soc. Conf. Comput. Vis. Pattern Recognit.*, vol. 2016-Decem, pp. 779–788, 2016, doi: 10.1109/CVPR.2016.91.
[19] W. Liu *et al.*, "Ssd: Single shot multibox detector," in *European conference on computer vision*, 2016, pp. 21–37.
[20] J. Redmon and A. Farhadi, "YOLO9000: Better, faster, stronger," *Proc. - 30th IEEE Conf. Comput. Vis. Pattern Recognition, CVPR 2017*, vol. 2017-Janua, pp. 6517–6525, 2017, doi: 10.1109/CVPR.2017.690.
[21] J. Redmon and A. Farhadi, "Yolov3: An incremental improvement," *arXiv Prepr. arXiv1804.02767*, 2018.
[22] A. Bochkovskiy, C. Y. Wang, and H. Y. M. Liao, "YOLOv4: Optimal Speed and Accuracy of Object Detection," *arXiv*, 2020.
[23] K. Liu, H. Tang, S. He, Q. Yu, Y. Xiong, and N. Wang, "Performance Validation of Yolo Variants for Object Detection," in *Proceedings of the 2021 International Conference on Bioinformatics and Intelligent Computing*, 2021, pp. 239–243.
[24] A. Zhu, Y. Zhu, N. Wang, and Y. Chen, "A robust self-driven surface crack detection algorithm using local features," *Insight Non-Destructive Test. Cond. Monit.*, vol. 62, no. 5, pp. 269–276, 2020, doi: 10.1784/insi.2020.62.5.269.
[25] S. He, H. Tang, X. Lu, H. Yan, and N. Wang, "MSHCNet: Multi-Stream Hybridized Convolutional Networks with Mixed Statistics in Euclidean/Non-Euclidean Spaces and Its Application to Hyperspectral Image Classification," *arXiv Prepr. arXiv2110.03346*, 2021.
[26] S. He *et al.*, "RSI-Net: Two-Stream Deep Neural Network Integrating GCN and Atrous CNN for Semantic Segmentation of High-resolution Remote Sensing Images," *arXiv Prepr. arXiv2109.09148*, 2021.
[27] M. D. Zeiler, D. Krishnan, G. W. Taylor, and R. Fergus, "Deconvolutional networks," in *2010 IEEE Computer Society Conference on computer vision and pattern recognition*, 2010, pp. 2528–2535.
[28] C. Xing, M. Wang, C. Dong, C. Duan, and Z. Wang, "Using Taylor expansion and convolutional sparse representation for image fusion," *Neurocomputing*, vol. 402, pp. 437–455, 2020.
[29] P. Bao *et al.*, "Convolutional sparse coding for compressed sensing CT reconstruction," *IEEE Trans. Med. Imaging*, vol. 38, no. 11, pp. 2607–2619, 2019.
[30] B. Wang, J. Deng, and Y. Sun, "Image super-resolution algorithm's research using convolutional sparse coding model," *Int. J. Inf. Commun. Technol.*, vol. 15, no. 1, pp. 92–106, 2019.
[31] V. Papyan, Y. Romano, J. Sulam, and M. Elad, "Convolutional dictionary learning via local processing," in *Proceedings of the IEEE International Conference on Computer Vision*, 2017, pp. 5296–5304.
[32] V. Papyan, J. Sulam, and M. Elad, "Working locally thinking globally: Theoretical guarantees for convolutional sparse coding," *IEEE Trans. Signal Process.*, vol. 65, no. 21, pp. 5687–5701, 2017.

> REPLACE THIS LINE WITH YOUR PAPER IDENTIFICATION NUMBER (DOUBLE-CLICK HERE TO EDIT) < 8[33] V. Papyan, Y. Romano, J. Sulam, and M. Elad, "Theoretical Foundations of Deep Learning via Sparse Representations."

[34] J. Sulam, V. Papyan, Y. Romano, and M. Elad, "Multilayer convolutional sparse modeling: Pursuit and dictionary learning," *IEEE Trans. Signal Process.*, vol. 66, no. 15, pp. 4090–4104, 2018, doi: 10.1109/TSP.2018.2846226.

[35] H. Tang *et al.*, "CSC-Unet: A Novel Convolutional Sparse Coding Strategy based Neural Network for Semantic Segmentation," *arXiv Prepr. arXiv2108.00408*, 2021.

[36] A. Aberdam, J. Sulam, and M. Elad, "Multi-layer sparse coding: The holistic way," *SIAM J. Math. Data Sci.*, vol. 1, no. 1, pp. 46–77, 2019.

[37] H. Tang *et al.*, "Comparison of Convolutional Sparse Coding Network and Convolutional Neural Network for Pavement Crack Classification: A Validation Study," *J. Phys. Conf. Ser.*, vol. 1682, no. 1, 2020, doi: 10.1088/1742-6596/1682/1/012016.

[38] V. Papyan, Y. Romano, and M. Elad, "Convolutional neural networks analyzed via convolutional sparse coding," *J. Mach. Learn. Res.*, vol. 18, pp. 1–52, 2017.

[39] J. Sulam, A. Aberdam, A. Beck, and M. Elad, "On multi-layer basis pursuit, efficient algorithms and convolutional neural networks," *IEEE Trans. Pattern Anal. Mach. Intell.*, vol. 42, no. 8, pp. 1968–1980, 2019.

[40] B. K. Natarajan, "Sparse approximate solutions to linear systems," *SIAM J. Comput.*, vol. 24, no. 2, pp. 227–234, 1995.

[41] E. J. Candès, J. Romberg, and T. Tao, "Robust uncertainty principles: Exact signal reconstruction from highly incomplete frequency information," *IEEE Trans. Inf. theory*, vol. 52, no. 2, pp. 489–509, 2006.

[42] M. Elad, *Sparse and redundant representations: From theory to applications in signal and image processing*. Springer New York, 2010.

[43] X. Glorot, A. Bordes, and Y. Bengio, "Deep sparse rectifier neural networks," in *Proceedings of the fourteenth international conference on artificial intelligence and statistics*, 2011, pp. 315–323.

[44] K. Bredies and D. A. Lorenz, "Linear convergence of iterative soft-thresholding," *J. Fourier Anal. Appl.*, vol. 14, no. 5–6, pp. 813–837, 2008.

[45] A. Paszke *et al.*, "PyTorch: An imperative style, high-performance deep learning library," *arXiv*, no. NeurIPS, 2019.

[46] A. Krizhevsky and G. Hinton, "Learning multiple layers of features from tiny images," 2009.

[47] Ç. F. Özgenel and A. G. Sorguç, "Performance comparison of pretrained convolutional neural networks on crack detection in buildings," in *ISARC. Proceedings of the International Symposium on Automation and Robotics in Construction*, 2018, vol. 35, pp. 1–8.